# THE APPLICATION OF ENERGY AND LAPLACIAN ENERGY OF HESITANCY FUZZY GRAPHS BASED ON SIMILARITY MEASURES IN DECISION-MAKING PROBLEMS

RAJAGOPAL REDDY N[1]

[1]*PhD Scholar, School of Advanced Sciences, Vellore Institute of Technology, Vellore 632 014, Tamilnadu, India*
*rajagopalreddy.n2019@vitstudent.ac.in*

S. SHARIEF BASHA[*]

[*]*Department of Mathematics, School of Advanced Sciences, Vellore Institute of Technology, Vellore 632 014, Tamilnadu, India.*
*shariefbasha.s@gmail.com*

*Abstract:* In this article, a new hesitancy fuzzy similarity measure is defined and then used to develop the matrix of hesitancy fuzzy similarity measures, which is subsequently used to classify hesitancy fuzzy graph using the working procedure. We build a working procedure (Algorithm) for estimating the eligible reputation scores values of experts by applying hesitancy fuzzy preference relationships (HFPRs) and the usual similarity degree of one distinct HFPRs to each other's. As the last step, we provide real time numerical examples to demonstrate and validate our working procedure.

## 1. Introduction

In the fuzzy set (FS) theory, fuzzy graphs are found. Zadeh [1]proposed a FS in 1965. Its aim was to establish an unclear and inaccurate theory of sets that also characterize the majority of real-life sets. Joseph Brown [2] was introduced the FSs including "holes," limited FSs and linked FSs. The concept of convex FSs and also derived some of the properties. He gave some of the conclusions on convex FS, and star-shaped FS. In 1966, Goguen [3] expanded the notion to include functions from a collection of the elements to a grid by introducing the FSs. Rosenfeld [4] studied fuzzy relations on fuzzy sets and established the layout of FGs, and he was able to derive analogues for a few graph conceptual concepts. And also he was the first person to developed FG theory. Fuzzy graphs (FGs) are found in fuzzy set theory. Binary relationships can be denoted as graphs. Fuzzy binary associations are denoted by graphs known as FGs. J.N. Mordeson and C.S. Peng [5] were introduced the idea of operations on FGs. And also implement the FG complements and few operations that can be performed on FGs. A few notes on FGs were made by Bhattacharya [6]. R. Parvathi and M.G. Karunambiga [7] have presented a new concept for IFG. Several IFG properties are taken into account and also examined using appropriate illustrations.

Anjali N, Sunil Mathew [8] was proposed the idea of energy is expanded to the FGs and also introduced the FG's adjacency matrix. The two limits and basic fundamentals of the energy of FGs are given. B. Praba and V.M. Chandrasekaran [9] was introduced the new idea of the energy of IFG and also given the intuitionistic fuzzy adjacency matrix (IFAM). They provide examples of the energy of IFG and obtained the two limitations of an IFG. An IFG's energy is developed from the idea of FG.S. Sharief Basha and E. Kartheek [10,11] was introduced the idea of Laplacian energy (LE) of IFG. In terms of its adjacency matrix, the IFAM and LE is defined. The two limitations of LE of IFG are derived and also provided several operations of LE of IFG. HFSs are recommended to deal with the frequent problem of determining the membership level in an object from several possibilities. Torra V [12] has brought more developments to the FSs and has dubbed its HFSs to assess lack of confidence. V Torra was established the idea of hesitancy fuzzy sets (HFSs) and several basic definitions are also defined. Zeshui Xu [13] defined the concept of HFSs theory and its applications. Also, the Hesitancy preference Relation(s) (HFPR) is derived. Vinoth kumar N and Geetharamani G [14]was introduced several operations in hesitancy fuzzy graphs (HFGs) and found properties on strong HFGs in various operations in HFGs. The newest graph named HFGs was established by T. Pathinathan, J. Jon Arockiaraj and J. Jesintha Rosline [15] and included several fundamental definitions and conceptual verification of HFGs. The notion for hesitation as a hesitant fuzzy entry that represents a decision-hesitance developer's in decision-making has been presented in [16–19] A variety of new distances and similarity measurements (SMs) are created among HFSs, considering both the HFE values and another values. The properties of SMs distance are also discussed. Obbu Ramesh and S. Sharief Basha were proposed the idea of cosine similarity in decision making problems (DMPs) by signless LE of IFGs. In 2018, Guiwu Wei and Yu Wei [20] were introduced the concept of the SMs between the Pythagorean fuzzy sets (PFSs), depending on the function, the membership grades, nonmemship grades and reluctance level in PFSs. Also they applied this grades to SMs and scores SMs of PFSs and illustrated the actual problems are provided.

The following is how the article is designed. Section 2 deals with the basic fundamentals related to HFG and certain measures of similarity between HFG. Section 3 offers a working procedure and to determine expert scores and ranks as regards a technique of SMs. Section 4 offers a procedure for using its "membership" and "nonmembership" scores. The effectiveness of predicted techniques through evaluations between other methodologies is demonstrated in implementation of DMPs.

**2. Preliminaries**

There are several basic fundamentals and properties related to HFG and certain measures of similarity of HFG that are provided here. We defined the following the properties of Energy of hesitancy fuzzy graphs and Laplacian energy of hesitancy fuzzy graph [35, 36]

**Definition 2.1:** Assuming $Y$ be a standard set, and an HFS on $Y$ is defined in terms of a function, when implemented to $Y$, gives a subset of [0,1], which may be written mathematically as the following symbol:

$$E = \{\langle y, h_E(Y)\rangle | y \epsilon Y|\}$$

Where $h_E(Y)$ is known as hesitant element and it is a collection of numbers in the range [0,1] indicating the degree to which the values $y \epsilon Y$ contains to the set E.

**Definition 2.2:** Suppose $HG$ be an $HFG$ and it is of the form $HG = (V, E, \mu, \gamma, \beta)$, where

(a) Consider $V = \{t_1, t_2, t_3 \ldots t_n\}$ such that $\mu_1 : V \to [0,1], \gamma_1 : V \to [0,1]$ and $\beta_1 : V \to [0,1]$ are denotes the grade of membership, nonmembership and hesitant of the elements $t_i \in V$ and $\mu_1(t_i) + \gamma_1(t_i) + \beta_1(t_i) = 1$, where

$$\beta_1(t_i) = 1 - [\mu_1(t_i) + \gamma_1(t_i)]$$

(b) Let $E \subseteq V \times V$ where $\mu_2 : V \times V \to [0,1]$, $\gamma_2 : V \times V \to [0,1]$ and $\beta_2 : V \times V \to [0,1]$ are such that,

$$\mu_2(t_i, t_j) \leq \min[\mu_1(t_i), \mu_1(t_j)]$$
$$\gamma_2(t_i, t_j) \leq \max[\gamma_1(t_i), \gamma_1(t_j)]$$
$$\beta_2(t_i, t_j) \leq \min[\beta_1(t_i), \beta_1(t_j)] \text{ and}$$
$$0 \leq \mu_2(t_i, t_j) + \gamma_2(t_i, t_j) + \beta_2(t_i, t_j) \leq 1, \forall (t_i, t_j) \in E.$$

**Definition 2.3:** Consider $HG = (V, E, \mu, \gamma, \beta)$ be a HFG and $\alpha_i, \delta_i$ and $\lambda_i$ are the set of Eigen values of an adjacency matrix $A(HG)$ of HFG then the energy of HFG is defined as

$$E(A(HG)) = E\left(A_\mu(HG), A_\gamma(HG), A_\beta(HG)\right) = \left(\sum_{i=1}^{n}|\alpha_i|, \sum_{i=1}^{n}|\delta_i|, \sum_{i=1}^{n}|\lambda_i|\right)$$

Where

$$\sum_{i=1}^{n}|\alpha_i|, \quad \sum_{i=1}^{n}|\delta_i|, \sum_{i=1}^{n}|\lambda_i|$$

is the energy of membership elements, nonmembership elements, hesitant elements, which are defined as

$$E\left(A_\mu(HG)\right), E\left(A_\gamma(HG)\right), E\left(A_\beta(HG)\right).$$

**Theorem 2.1:** Let $HG$ be an HFG (not having the loops) with $|V| = p$ and $|E| = q$ and $A(HG) = \left((\mu_{ij})(\gamma_{ij})(\beta_{ij})\right)$ be Hesitancy fuzzy adjacency matrix of $G$ then

(i) $\sqrt{p(p-1)|A|^{\frac{2}{p}} + 2\sum_{1 \leq i \leq j \leq p} \mu_{ij}\mu_{ji}} \leq E\left(\mu_{ij}(HG)\right) \leq \sqrt{2p \sum_{1 \leq i \leq j \leq p} \mu_{ij}\mu_{ji}}$

Where $|A|$ is the determinant of $A\left(\mu_{ij}(HG)\right)$

$$(ii)\ \sqrt{p(p-1)|B|^{\frac{2}{p}} + 2\sum_{1\leq i\leq j\leq p}\gamma_{ij}\gamma_{ji}} \leq E\left(\mu_{ij}(HG)\right) \leq \sqrt{2p\sum_{1\leq i\leq j\leq p}\gamma_{ij}\gamma_{ji}}$$

Where $|B|$ is the determinant of $B\left(\mu_{ij}(HG)\right)$

$$(iii)\ \sqrt{p(p-1)|C|^{\frac{2}{p}} + 2\sum_{1\leq i\leq j\leq p}\beta_{ij}\beta_{ji}} \leq E\left(\mu_{ij}(HG)\right) \leq \sqrt{2p\sum_{1\leq i\leq j\leq p}\beta_{ij}\beta_{ji}}$$

Where $|C|$ is the determinant of $C\left(\mu_{ij}(HG)\right)$

**Theorem 2.2:** Let $HG$ be the HFG with $|V| = p$ and $|V| = q$. Let $A$, $B$, $C$ are adjacency matrices then

$$(i)\ E\left(\mu_{ij}(HG)\right) \leq \frac{2\sum_{i=1}^{q}\mu_i^2}{p} + \sqrt{(p-1)\left\{2\sum_{i=1}^{q}\mu_i^2 - \left(\frac{2\sum_{i=1}^{q}\mu_i^2}{p}\right)^2\right\}}$$

$$(ii)\ E\left(\gamma_{ij}(HG)\right) \leq \frac{2\sum_{i=1}^{q}\gamma_i^2}{p} + \sqrt{(p-1)\left\{2\sum_{i=1}^{q}\gamma_i^2 - \left(\frac{2\sum_{i=1}^{q}\gamma_i^2}{p}\right)^2\right\}}\ \text{and}$$

$$(iii)\ E\left(\beta_{ij}(HG)\right) \leq \frac{2\sum_{i=1}^{q}\beta_i^2}{p} + \sqrt{(p-1)\left\{2\sum_{i=1}^{q}\beta_i^2 - \left(\frac{2\sum_{i=1}^{q}\beta_i^2}{p}\right)^2\right\}}$$

**Definition 2.4:** Let $HG = (V, E, \mu, \gamma, \beta)$ is a HFG and then the Laplacian matrix of a HFG is described as

$$L(HG) = \left(L_\mu(HG), L_\gamma(HG), L_\beta(HG)\right),$$

Where $L_\mu(HG) = D_\mu(HG) - A_\mu(HG)$, $L_\gamma(HG) = D_\gamma(HG) - A_\gamma(HG)$ and $L_\beta(HG) = D_\beta(HG) - A_\beta(HG)$. Here $D_\mu(HG)$ be the degree matrix of the membership, $D_\gamma(HG)$ be the degree matrix of the nonmembership $D_\beta(HG)$ is the degree matrix of the hesitant elements and $A_\mu(HG)$ is the membership element matrix, $A_\gamma(HG)$ is the nonmembership element matrix, $A_\beta(HG)$ is the hesitant element matrix.

**Definition 2.5:** Let $HG = (V, E, \mu, \gamma, \beta)$ is a HFG and then the Laplacian energy of a HFG is described as

$$LE(HG) = \left(LE_\mu(HG), LE_\gamma(HG), LE_\beta(HG)\right)$$

Where

$$LE_\mu(HG) = \sum_{i=1}^{n}\left|\lambda_i - \frac{2\sum_{1\leq i\leq j\leq n}(\mu_{ij} + \mu_{ji})}{n}\right|$$

be the Laplacian energy of membership value,

$$LE_\gamma(HG) = \sum_{i=1}^{n} \left| \alpha_i - \frac{2\sum_{1\leq i\leq j\leq n}(\gamma_{ij} + \gamma_{ji})}{n} \right|$$

be the Laplacian energy of nonmembership value, and

$$LE_\beta(HG) = \sum_{i=1}^{n} \left| \delta_i - \frac{2\sum_{1\leq i\leq j\leq n}(\beta_{ij} + \beta_{ji})}{n} \right|$$

be the Laplacian energy of hesitant value

**Theorem 2.3:-** Let $HG = (V, E, \mu, \gamma, \beta)$ be an HFG with $|V| = n$ vertices and $\lambda_1 \geq \lambda_2 \geq \cdots \geq \lambda_n$ is the Laplacian eigenvalues of membership values of Hesitancy fuzzy graph then

$$(a) \sum_{i=1}^{n} \lambda_i^2 = 2 \sum_{1\leq i\leq j\leq n} \mu_{ij} \qquad (b) \sum_{i=1}^{n} \lambda_i^2 = 2 \sum_{1\leq i\leq j\leq n} \mu_{ij}^2 + \sum_{i=1}^{n} d_{\mu_{ij}}^2(v_i)$$

**Corollary 2.4:** Let $HG = (V, E, \mu, \gamma, \beta)$ be a HFG of membership function with $|V| = n$ vertices and $\lambda_1 \geq \lambda_2 \geq \cdots \geq \lambda_n$ is the membership Laplacian eigenvalues of Hesitancy fuzzy graph $HG$, where $\Psi_i = \lambda_i - \frac{2\sum_{1\leq i\leq j\leq n} \mu_{ij}^2}{n}$ then we obtain

$$(a) \sum_{i=1}^{n} \Psi_i = 0 \qquad (b) \sum_{i=1}^{n} \Psi_i^2 = 2M$$

Where

$$M = \sum_{1\leq i\leq j\leq n} \mu_{ij}^2 + \frac{1}{2}\sum_{i=1}^{n} \left( d_{\mu_{ij}(\check{G})}(v_i) - \frac{2\sum_{1\leq i\leq j\leq n} \mu_{ij}}{n} \right)^2$$

Similarly for membership Laplacian eigenvalues of HFG, where $\chi_i = \delta_i - \frac{2\sum_{1\leq i\leq j\leq n} \gamma_{ij}}{n}$ then we have

$$(a) \sum_{i=1}^{n} \chi_i = 0 \qquad (b) \sum_{i=1}^{n} \chi_i^2 = 2N$$

Where

$$N = \sum_{1\leq i\leq j\leq n} \gamma_{ij}^2 + \frac{1}{2}\sum_{i=1}^{n} \left( d_{\gamma_{ij}(\check{G})}(v_i) - \frac{2\sum_{1\leq i\leq j\leq n} \gamma_{ij}}{n} \right)^2$$

In the same way, for Hesitancy of the element Laplacian eigenvalues of Hesitancy fuzzy graph $HG$, where $\Psi_i = \lambda_i - \frac{2\sum_{1\leq i\leq j\leq n} \mu_{ij}^2}{n}$ then we obtain

$$(a) \sum_{i=1}^{n} \Psi_i = 0 \qquad (b) \sum_{i=1}^{n} \Psi_i^2 = 2R$$

Where $R = \sum_{1\leq i\leq j\leq n} \beta_{ij}^2 + \frac{1}{2}\sum_{i=1}^{n} \left( d_{\beta_{ij}(\check{G})}(v_i) - \frac{2\sum_{1\leq i\leq j\leq n} \beta_{ij}}{n} \right)^2$

**Definition 2.6:** Let $HG = (V, E, \mu, \gamma, \beta)$ be HFG with $|V| = n$ vertices and $\lambda_1 \geq \lambda_2 \geq \cdots \geq \lambda_n$ are the membership eigenvalues of Laplacian matrix $HG$. The Laplacian energy of Hesitancy fuzzy graph $HG$ is defined as,

$$LE(\mu_{ij}(HG)) = \left|\lambda_i - \frac{2\sum_{1\leq i\leq j\leq n}\mu(v_i, v_j)}{n}\right|$$

The Laplacian energy of Hesitancy fuzzy graph $HG = (V, E, \mu, \gamma, \beta)$ is defined as

$$[LE(\mu_{ij}(HG)), LE(\gamma_{ij}(HG)), LE(\beta_{ij}(HG))]$$

**Theorem 2.5:** Let $HG = (V, E, \mu, \gamma, \beta)$ be an HFG (without loops) with $|V| = n$ and $|E| = m$ and $HG = ((\mu_{ij}), (\gamma_{ij}), (\beta_{ij}))$ be a Hesitancy fuzzy adjacency matrix and $L(HG) = (L(\mu_{ij}), L(\gamma_{ij}), L(\beta_{ij}))$ be a Laplacian of Hesitancy fuzzy matrix of $\breve{G}$ then

$(i)\ LE(\mu_{ij}(HG)) \leq \sqrt{2n\left(\frac{1}{2}\sum_{i=1}^{n}\left(d_{\mu_{ij}(\breve{G})}(v_i) - \frac{2\sum_{1\leq i\leq j\leq n}\mu_{ij}}{n}\right)^2 + \sum_{1\leq i\leq j\leq n}\mu_{ij}^2\right)}$

$(ii)\ LE(\gamma_{ij}(HG)) \leq \sqrt{2n\left(\frac{1}{2}\sum_{i=1}^{n}\left(d_{\gamma_{ij}(\breve{G})}(v_i) - \frac{2\sum_{1\leq i\leq j\leq n}\gamma_{ij}}{n}\right)^2 + \sum_{1\leq i\leq j\leq n}\gamma_{ij}^2\right)}$

$(iii)\ LE(\beta_{ij}(HG)) \leq \sqrt{2n\left(\frac{1}{2}\sum_{i=1}^{n}\left(d_{\beta_{ij}(\breve{G})}(v_i) - \frac{2\sum_{1\leq i\leq j\leq n}\beta_{ij}}{n}\right)^2 + \sum_{1\leq i\leq j\leq n}\beta_{ij}^2\right)}$

**Theorem 2.6:** Let $HG = (V, E, \mu, \gamma, \beta)$ be a Hesitancy fuzzy graph with vertices $|V| = n$ and $L(HG) = (L(\mu_{ij}), L(\gamma_{ij}), L(\beta_{ij}))$ be Laplacian Hesitancy fuzzy matrix of $HG$ then

$(i)\ LE(\mu_{ij}(HG)) \geq \sqrt{2\left(\sum_{i=1}^{n}\left(d_{\mu_{ij}(\breve{G})}(v_i) - \frac{2\sum_{1\leq i\leq j\leq n}\mu_{ij}}{n}\right)^2 + 2\sum_{1\leq i\leq j\leq n}\mu_{ij}^2\right)}$

$(ii)\ LE(\gamma_{ij}(HG)) \geq \sqrt{2\left(\sum_{i=1}^{n}\left(d_{\gamma_{ij}(\breve{G})}(v_i) - \frac{2\sum_{1\leq i\leq j\leq n}\gamma_{ij}}{n}\right)^2 + 2\sum_{1\leq i\leq j\leq n}\gamma_{ij}^2\right)}$

$(iii)\ LE(\beta_{ij}(HG)) \geq \sqrt{2\left(\sum_{i=1}^{n}\left(d_{\beta_{ij}(\breve{G})}(v_i) - \frac{2\sum_{1\leq i\leq j\leq n}\beta_{ij}}{n}\right)^2 + 2\sum_{1\leq i\leq j\leq n}\beta_{ij}^2\right)}$

**Theorem 2.7:** Let $HG = (V, E, \mu, \gamma, \beta)$ be a Hesitancy fuzzy graph with vertices $|V| = n$ and the Laplacian Hesitancy fuzzy matrix of $HGL(HG) = (L(\mu_{ij}), L(\gamma_{ij}), L(\beta_{ij}))$ then

(i) $LE(\mu_{ij}(HG)) \leq$

$$\Psi_1 + \sqrt{(n-1)\left(\sum_{i=1}^{n}\left(d_{\mu_{ij}(\breve{G})}(v_i) - \frac{2\sum_{1\leq i\leq j\leq n}\mu_{ij}}{n}\right)^2 + 2\left(\sum_{1\leq i\leq j\leq n}\mu_{ij}^2\right) - \Psi_1^2\right)}$$

(ii) $LE(\gamma_{ij}(HG)) \leq$

$$\Psi_1 + \sqrt{(n-1)\left(\sum_{i=1}^{n}\left(d_{\gamma_{ij}(\breve{G})}(v_i) - \frac{2\sum_{1\leq i\leq j\leq n}\gamma_{ij}}{n}\right)^2 + 2\left(\sum_{1\leq i\leq j\leq n}\gamma_{ij}^2\right) - \Psi_1^2\right)}$$

(iii) $LE(\beta_{ij}(HG)) \leq$

$$\Psi_1 + \sqrt{(n-1)\left(\sum_{i=1}^{n}\left(d_{\beta_{ij}(\breve{G})}(v_i) - \frac{2\sum_{1\leq i\leq j\leq n}\gamma_{ij}}{n}\right)^2 + 2\left(\sum_{1\leq i\leq j\leq n}\beta_{ij}^2\right) - \Psi_1^2\right)}$$

**Definition 2.7: Hesitancy Fuzzy Preference Relations**

Expert is usually obligatory to give its tendencies to replacements throughout the GDM procedure. The expert may make judgments with confidence, but the experts are not always assured of them. And that it is appropriate to stimulate the preference norms of the expert with Hesitancy fuzzy numbers considerably above the analytical norms. Following that, we discussed how to acquire far more information from the experts' preferences over the replacements throughout the order to change the provided resting scores of experts for most reasonable group decision making.

**Definition 2.8:** Consider $R$ be an HFPR on the set $X$ is represented by a matrix $R = (a_{ij})_{m\times m}$ with $a_{ij} = (\mu_{ij}, \gamma_{ij}, \beta_{ij}, \pi_{ij})$, where $i, j$ are having $1, 2, \ldots, n$ and the hesitancy fuzzy value is $a_{ij}$, consisting of the degree of confidence of $\mu_{ij}$ to which $x_i$ is preferable to $x_j$, the degree of confidence $\gamma_{ij}$ to which $x_i$ is non-preferable to $x_j$, $\beta_{ij}$ to which $x_i$ is preferred to $x_j$ and the degree of confidence $\pi_{ij}$ of which $x_i$ is preferred to $x_j$. Further, $\mu_{ij}, \gamma_{ij}, \beta_{ij}$, and $\pi_{ij}$ have the following properties:

(i) $0 \leq \mu_{ij} + \gamma_{ij} + \beta_{ij} \leq 1$,

(ii) $\pi_{ij} = 1 - \mu_{ij} - \gamma_{ij} - \beta_{ij}$,

(iii) $\mu_{ii} = \gamma_{ii} = \beta_{ii} = 0$, and $\pi_{ij} = 0$, $i, j$ is having $1, 2, \ldots, n$

**Theorem 2.8:** Consider $R^{(p)} = (a_{ij}^{(p)})_{m\times m}$ be HFPR given by the experts $e_i$ where $i$ and $p$ are having $1, 2, \ldots, n$ respectively, and $\omega = \omega_1, \omega_2, \ldots, \omega_p$ be the weight vector of experts, where $\omega_p \geq 0$ and $\sum_{p=1}^{n}\omega_p = 1$, then the aggregation $R = (a_{ij})_{m\times m}$ of $R^{(p)} = (a_{ij}^{(p)})_{m\times m}$ ($p$ is having $1, 2, \ldots, n$) is also an HFPR, such that

$$a_{ij} = (\mu_{ij}, \gamma_{ij}, \beta_{ij}, \pi_{ij})$$
$$\mu_{ij} = \sum_{p=1}^{n}\omega_p \mu_{ij}^{(p)}, \gamma_{ij} = \sum_{p=1}^{n}\omega_p \gamma_{ij}^{(p)}, \beta_{ij} = \sum_{p=1}^{n}\omega_p \beta_{ij}^{(p)} \text{ and}$$
$$\pi_{ij} = \sum_{p=1}^{n}\omega_p \pi_{ij}^{(p)}, \mu_{ii} = \gamma_{ii} = \beta_{ii} = \pi_{ii} = 0 \text{ } (i, j \text{ are having } 1, 2, \ldots, n).$$

**Definition 2.9:** Assume that the HFPR be $R^{(p)} = \left(a_{ij}^{(p)}\right)_{m \times m}$ (where $p$ is having $1, 2, \ldots, n$) be $n$ HFPRs and $R = \left(a_{ij}\right)_{m \times m}$ of $R^{(p)} = \left(a_{ij}^{(p)}\right)_{m \times m}$ ($p$ is having $1, 2, \ldots, n$) represent their collected HFPR, then the similarity measure between $R^{(p)}$ and $R$ is

$$S(R^{(p)}, R) = \frac{1}{n}\sum_{i=1}^{n} S\left(a_{ij}^{(p)}, a_{ij}\right)$$

Where $S\left(a_{ij}^{(p)}, a_{ij}\right)$ is the degree of similarity measure between $a_{ij}^{(p)}$ and $a_{ij}$.

The degree of similarity between the individual hesitancy fuzzy preference relation $R^{(p)}$ and the aggregate $R$ is reflected in the similarity measure $S(R^{(p)}, R)$. In several real life scenarios, some specialists may provide excessively high or excessively low priority assertions for their favoured or reprehensible items, resulting in low levels of agreement between the collective hesitancy fuzzy preference relation and the individual hesitancy fuzzy preference relations. In these kinds of circumstances, we would give these specialists minimal weights in the decision-making approach. The higher degree of the similarity measure $S(R^{(p)}, R)$, the greater the weight of the specialist $e_i$. As a consequence, we suggest the following formula for calculating specialist weights:

$$\omega_p = \frac{S(R^{(p)}, R)}{\sum_{p=1}^{n} S(R^{(p)}, R)}$$

Where $p$ is having $1, 2, \ldots, n$.

**Definition 2.10:** If $P$ and $R$ are HFGs then the similarity measures from $P$ to $R$ be denoted as $S(P, R)$, it has the following characteristics:

i) $0 \leq S(P, R) \leq 1$; $\hspace{5em}$ (S1)
ii) $S(P, R) = 1$, if $f\ A = B$; $\hspace{3em}$ (S2)
iii) $S(P, R) = S(R, P)$; $\hspace{5em}$ (S3)
iv) If $P \subseteq R \subseteq B$, then $S(P, B) \leq S(P, R)$ and $S(P, B) \leq S(R, B)$. $\hspace{1em}$ (S4)

## 2.1. *A Technique for Determining Experts Scores*

Let $X$ be the replacements set, and the experts set $E$. The expert offers proof of preference to every replacement and establishes HFPRs

$$A^l = \left(a_{ij}^l\right)_{m \times m}$$

Where $a_{ij}^l = \left(p_{ij}^l, r_{ij}^l, q_{ij}^l\right)$ and $0 \leq p_{ij}^l + r_{ij}^l + q_{ij}^l \leq 1$.

The energy and Laplacian energy can determine the ambiguous HFGs indicator. Every HFPR $A^l$ seems to be an HFG in the adjacent matrix such that the ambiguous specify is determined by energy and the measurement of Laplacian energy. We typically think of the degree of ambiguity of intuitional tendency as unlikely to provide more trust to degrees achieved through the GDM technique. We developed the following Technique to score the 'impartial' burdens of the experts.

## 2.2. *Flow Chart*

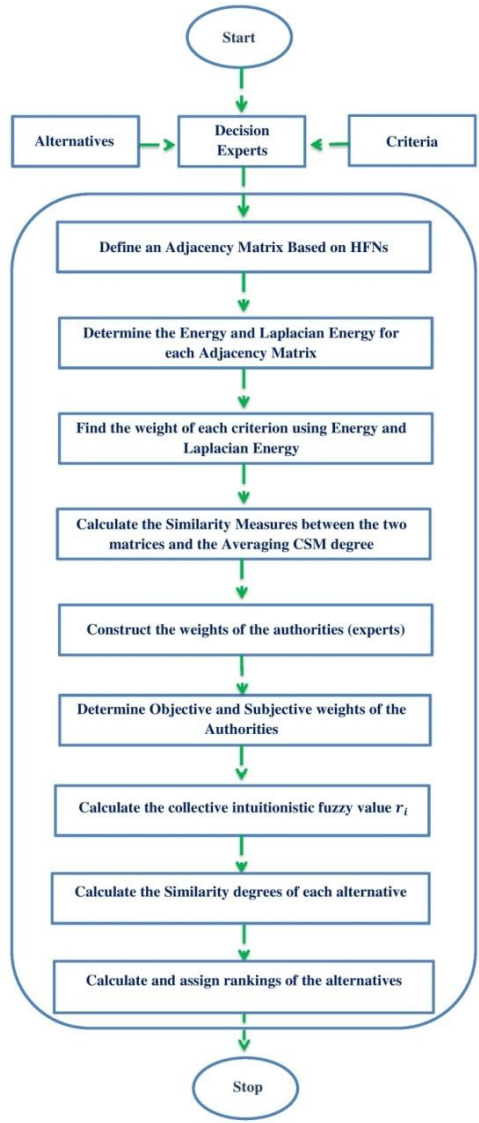

## 2.3. *Working Procedure*

In this subsection, working procedure is constructed for GDM actual issues focused on HFPRs.

We describe $C = c_1, c_2, c_3, \ldots, c_m$ as an impartial scoresing vector of expert for GDM issues based on HFPRs, where $C_b > 0$, $b = 1,2,3,\ldots,l$, and the total of all the scoring values of the experts is equal to one is written as $\sum_{i=1}^{l} C_i = 1$.

**Stage i.**

(a) Evaluate the energy $E(M^{(b)})$ of $M^{(b)}$:
$$E(M^{(k)}) = |\sum_{i=1}^{n} \kappa_i| \tag{2.1}$$

(b) Evaluate the Laplacian energy $LE(M^{(k)})$ of $M^{(k)}$:
$$LE(M^{(k)}) = \left|\kappa_i - \frac{2\sum_{1\leq i\leq j\leq n}\mu(t_i,t_j)}{n}\right| \tag{2.2}$$

**Stage ii.**

(a) Evaluate the scores $C_b^{\,1}$, determined by $E(M^{(k)})$, of the expert $e_b$:
$$C_b^{\,1} = \left((C_\mu)_{i'}, (C_\gamma)_{i'}, (C_\beta)_i\right) = \left[\frac{E((D_\mu)_i)}{\sum_{r=1}^{l} E((D_\mu)_r)}, \frac{E((D_\gamma)_i)}{\sum_{r=1}^{l} E((D_\gamma)_r)}, \frac{E((D_\beta)_i)}{\sum_{l=1}^{l} E((D_\beta)_r)}\right] \tag{2.3}$$

(b) Evaluate the scores $C_b^{\,1}$, determined by $LE(M^{(k)})$, of the expert $e_b$:
$$C_b^{\,1} = \left((C_\mu)_{i'}, (C_\gamma)_{i'}, (C_\beta)_i\right) = \left[\frac{LE((D_\mu)_i)}{\sum_{r=1}^{l} LE((D_\mu)_r)}, \frac{LE((D_\gamma)_i)}{\sum_{r=1}^{l} LE((D_\gamma)_r)}, \frac{LE((D_\beta)_i)}{\sum_{l=1}^{l} LE((D_\beta)_r)}\right] \tag{2.4}$$

**Stage iii.** Evaluate the SM $S(M^{(b)}, M^{(d)})$ between $M^{(b)}$ and $M^{(d)}$ for every $b \neq d$
$$S(M^{(b)}, M^{(d)}) = \frac{1}{n} + \frac{2}{n^2}\sum_{i=1}^{n}\sum_{j=i+1}^{n}\frac{1-\min\{|\mu_{ij}^b-\mu_{ij}^d|,|\gamma_{ij}^b-\gamma_{ij}^d|,|\beta_{ij}^b-\beta_{ij}^d|\}}{1+\max\{|\mu_{ij}^b-\mu_{ij}^d|,|\gamma_{ij}^b-\gamma_{ij}^d|,|\beta_{ij}^k-\beta_{ij}^d|\}} \tag{2.5}$$

Then the average similarity degree $S(M^{(b)})$ of $M^{(b)}$ to the others is calculated by
$$S(M^{(b)}) = \frac{1}{m-1}\sum_{l=1,b\neq d}^{n} S(M^{(b)}, M^{(d)}), b = 1,2,3,\ldots,l \tag{2.6}$$

**Stage iv.** Evaluate the scores $C_b^{\,a}$, determined by $S(M^{(b)})$ of the expert $e_b$:
$$C_b^{\,a} = \frac{S(M^{(b)})}{\sum_{i=1}^{l} S(M^{(i)})}, b = 1,2,3,\ldots,l \tag{2.7}$$

**Stage v.** Evaluate the "objective" scores $C_b^{\,2}$ of the expert $e_b$
$$C_b^{\,2} = \eta\, C_b^{\,1} + (1-\eta)\, C_b^{\,a}, \forall \eta \epsilon [0,1], b = 1,2,3,\ldots,l \tag{2.8}$$

**Stage vi.** Evaluate the subjective and objective scores $C_b^{\,1}$ and $C_b^{\,2}$ of the expert $e_b$
$$C_b = \gamma\, C_b^{\,1} + (1-\gamma)\, C_b^{\,2}, \forall \gamma \epsilon [0,1], b = 1,2,3,\ldots,l \tag{2.9}$$

**Stage vii.** Evaluate the cooperative HFPR $M = (r_{ij})_{n \times n}$ by

$$r_{ij} = (\mu_{ij}, \gamma_{ij}, \beta_{ij}) = \left(\sum_{b=1}^{l} C_b \mu_{ij}^{(b)}, \sum_{b=1}^{l} C_b \gamma_{ij}^{(b)}, \sum_{b=1}^{l} C_b \beta_{ij}^{(b)}\right) \quad (2.10)$$

for all $i, j = 1,2,3, \ldots, n$.

**Stage viii.**

(a) To evaluate the SMs $S(M^i, M^+)$ between $M^i$ and $M^+$ for every replacement $t_i$

$$S(M^{(i)}, M^{(+)}) = \frac{1}{n}\sum_{j=i}^{n} \frac{1 - min\{|\mu_{ij} - 1|, |\gamma_{ij} - 0|, |\beta_{ij} - 1|\}}{1 + max\{|\mu_{ij} - 1|, |\gamma_{ij} - 0|, |\beta_{ij} - 1|\}}$$

$$= \frac{1}{n}\sum_{j=i}^{n} \frac{1 - min\{1 - \mu_{ij}, \gamma_{ij}, 1 - \beta_{ij}\}}{1 + max\{1 - \mu_{ij}, \gamma_{ij}, 1 - \beta_{ij}\}} \quad (2.11)$$

(b) To evaluate the SMs $S(M^i, M^-)$ between $M^i$ and $M^-$ for every replacement $t_i$

$$S(M^{(i)}, M^{(-)}) = \frac{1}{n}\sum_{j=i}^{n} \frac{1 - min\{|\mu_{ij} - 0|, |\gamma_{ij} - 1|, |\beta_{ij} - 0|\}}{1 + max\{|\mu_{ij} - 0|, |\gamma_{ij} - 1|, |\beta_{ij} - 0|\}}$$

$$= \frac{1}{n}\sum_{j=i}^{n} \frac{1 - min\{\mu_{ij}, 1 - \gamma_{ij}, \beta_{ij}\}}{1 + max\{\mu_{ij}, 1 - \gamma_{ij}, \beta_{ij}\}} \quad (2.12)$$

**Stage ix.** Evaluate the values of $f(t_i)$, for every replacement $t_i$

$$f(t_i) = \frac{S(M^{(i)}, M^{(+)})}{S(M^{(i)}, M^{(+)}) + S(M^{(i)}, M^{(-)})} \quad (2.13)$$

The highest value of $f(t_i)$ is greater to the replacements $t_i$.

Now the order of ranking of the replacements is conformed.

### 2.4. *Application: Selection of finest Smart Phone Model*

Smartphones are among the most popular digital gadgets, accounting for a major fraction of our everyday life. There are several smartphones that have made our lives more comfortable and simple. Mr. Punarv intends to purchase an Apple smartphone for a variety of purposes. In the Apple firm, there are a variety of models available for the purchase of smartphones. However, he requires the selection of the finest model for acquiring the sophisticated features, storage, variant and design smartphones available in the marketplace from Apple. Suppose that $X = \{A_1, A_2, A_3, A_4\}$e the set of alternatives (we assigning the Four Apple smartphone models are $A_1 = iPhone\ 13\ Pro$, $A_2 = iPhone\ SE$, $A_3 = iPhone\ 12\ Pro\ max$, and $A_2 = iPhone\ 12\ mini$) be the set of four Apple firm smartphone models, and also assign three experts $e_b$, (where $i$ is having 1, 2, 3) for the selection of the finest smartphone model. $T = \{t_1, t_2, t_3\}$ be the set of criteria (parameters) which indicatessophisticated features, storage, design and variant respectively. Mr. Punarv evaluates the four smartphone models $A_i$, (where $i$ is having $1, 2, 3, 4$) from the Apple firm. The selection based on the criteria's features, storage, design and variant and offers its preference relations in the form of HFPR $R = (a_{ij})_{m \times m}$, where $a_{ij} = (\mu_{ij}, \gamma_{ij}, \beta_{ij})$ is the hesitancy fuzzy value allocated by Mr. Punarv expert with $\mu_{ij}$ the degree to which the model $A_i$ regarding the specified criteria, $\gamma_{ij}$ as a degree to which the model $A_i$ regarding the specified criteria and $\beta_{ij}$ as a degree to which the model $A_i$ regarding the specified criteria. Assume that the scores for every expert are $0, 0.2, 0.3, 0.5, 0.7$ and $1.0$ respectively and builds the HFPRs $R = (a_{ij})_{m \times m}$ for the specified criteria is represented in the matrix as shown below.

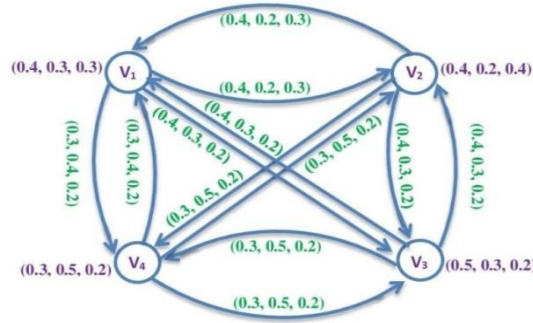

**Figure 1.** HFPR for criteria $t_1$

An adjacency Matrix from figure-1

$$M^{(1)} = M(HG) = \begin{bmatrix} (0,0,0) & (0.4, 0.2, 0.3) & (0.4, 0.3, 0.2) & (0.3, 0.4, 0.2) \\ (0.4, 0.2, 0.3) & (0,0,0) & (0.4, 0.3, 0.2) & (0.3, 0.5, 0.2) \\ (0.4, 0.3, 0.2) & (0.4, 0.3, 0.2) & (0,0,0) & (0.3, 0.5, 0.2) \\ (0.3, 0.4, 0.2) & (0.3, 0.5, 0.2) & (0.3, 0.5, 0.2) & (0,0,0) \end{bmatrix}$$

The energy of the matrix $M^{(1)}$ of HFG is

$$E(M^{(1)}) = (2.1114, 2.2436, 1.3062)$$

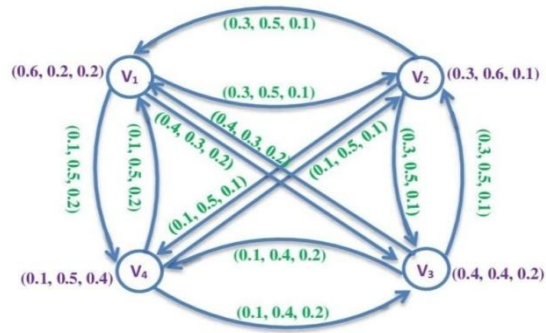

**Figure 2.** HFPR for criteria $t_2$

An adjacency Matrix from figure-2

$$M^{(2)} = M(HG) = \begin{bmatrix} (0,0,0) & (0.3,0.5,0.1) & (0.4,0.3,0.2) & (0.1,0.5,0.2) \\ (0.3,0.5,0.1) & (0,0,0) & (0.3,0.5,0.1) & (0.1,0.5,0.1) \\ (0.4,0.3,0.2) & (0.3,0.5,0.1) & (0,0,0) & (0.1,0.4,0.2) \\ (0.1,0.5,0.2) & (0.1,0.5,0.1) & (0.1,0.4,0.2) & (0,0,0) \end{bmatrix}$$

The energy of the matrix $M^{(2)}$ of $HFG$ is
$$E(M^{(2)}) = (1.4223, 2.7133, 0.9292)$$

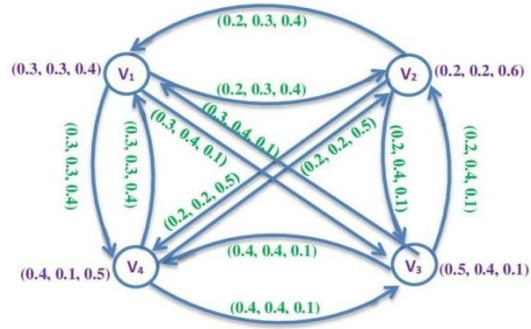

**Figure 3.** HFPR for criteria $t_3$

An adjacency Matrix from figure-3
$$M^{(3)} = M(HG) = \begin{bmatrix} (0,0,0) & (0.2,0.3,0.4) & (0.3,0.4,0.1) & (0.3,0.3,0.4) \\ (0.2,0.3,0.4) & (0,0,0) & (0.2,0.4,0.1) & (0.2,0.2,0.5) \\ (0.3,0.4,0.1) & (0.2,0.4,0.1) & (0,0,0) & (0.4,0.4,0.1) \\ (0.3,0.3,0.4) & (0.2,0.2,0.5) & (0.4,0.4,0.1) & (0,0,0) \end{bmatrix}$$

The energy of the matrix $M^{(3)}$ of $HFG$ is
$$E(M^{(3)}) = (1.6317, 2.0204, 1.8034)$$

*(i) The energy of Hesitancy Fuzzy Graph*

The energy of the adjacency matrices $M^{(1)}, M^{(2)}$ and $M^{(3)}$ of $HFG$ are
$$E(M^{(1)}) = (2.1114, 2.2436, 1.3062),$$
$$E(M^{(2)}) = (1.4223, 2.7333, 0.9292)$$
$$E(M^{(3)}) = (1.6317, 2.0204, 1.8034)$$

The scores of each expert can calculate as:
$$C_i^1 = \left((C_\mu)_i, (C_\gamma)_i, (C_\beta)_i\right) = \left[\frac{E\left((D_\mu)_i\right)}{\sum_{r=1}^t E\left((D_\mu)_t\right)}, \frac{E\left((D_\gamma)_i\right)}{\sum_{r=1}^t E\left((D_\gamma)_t\right)}, \frac{E\left((D_\beta)_i\right)}{\sum_{r=1}^t E\left((D_\beta)_t\right)}\right]$$

for $i = 1,2,3,\dots,n$

To evaluate the scores of every expert is as follows
$$C_1^1 = [0.3730, 0.3963, 0.2307],$$
$$C_2^1 = [0.2808, 0.5357, 0.1835],$$
$$C_3^1 = [0.2991, 0.3703, 0.3306]$$

To evaluate the SM $S(M^{(b)}, M^{(d)})$ between $M^{(b)}$ and $M^{(d)}$ using (5), we get,

$S(M^{(1)}, M^{(2)}) = 1.9856,$   $S(M^{(2)}, M^{(3)}) = 2.0579,$   $S(M^{(1)}, M^{(3)}) = 1.9155.$

The mean similarity degree (SD) $S(M^{(b)})$ of $M^{(b)}$ is obtained as below,

$S(M^{(1)}) = 1.9506,$   $S(M^{(2)}) = 2.0213,$   $S(M^{(3)}) = 1.9863.$

By calculating the values of the scores $C_b{}^a$ using (7), we get

$$C^b = (0.3274, 0.3392, 0.3334)$$

Using the formulae (7), evaluate an objective scores $C_b{}^2$ using $\eta = 0.5$, we get

$$C_1{}^2 = [0.3502, 0.3678, 0.2821],$$
$$C_2{}^2 = [0.3041, 0.4375, 0.2584]$$
$$C_3{}^2 = [0.3133, 0.3548, 0.4161].$$

Now determine the values of the scores $C_b$ using $\gamma = 0.5$, objective and subjective scores, we have

$$C_1 = [0.3616, 0.3821, 0.2564],$$
$$C_2 = [0.2925, 0.4866, 0.2692]$$
$$C_3 = [0.3062, 0.3626, 0.3734].$$

To determine the hesitancy preference relation $M = (r_{ij})_{n \times n}$, we get

$$M = \begin{bmatrix} (0,0,0) & (0.2936, 0.4285, 0.2532) & (0.3535, 0.4057, 0.1425) & (0.2296, 0.0.5049, 0.2545) \\ (0.2936, 0.4285, 0.2532) & (0,0,0) & (0.2936, 0.5030, 0.1155) & (0.1990, 0.5069, 0.2649) \\ (0.3535, 0.4057, 0.1425) & (0.2936, 0.5030, 0.1155) & (0,0,0) & (0.2602, 0.5307, 0.1425) \\ (0.2296, 0.0.5049, 0.2545) & (0.1990, 0.5069, 0.2649) & (0.2602, 0.5307, 0.1425) & (0,0,0) \end{bmatrix}$$

By evaluating the 'SMs' $S(M^i, M^+)$ between $M^i$ and $M^+$ for every replacement, we have

$S(M^1, M^+) = 0.3567,$   $S(M^2, M^+) = 0.3438,$
$S(M^3, M^+) = 0.3341,$   $S(M^4, M^+) = 0.3292$

By evaluating the 'SMs' $S(M^i, M^-)$ between $M^i$ and $M^-$ for every replacement, we have

$S(M^1, M^-) = 0.5071,$   $S(M^2, M^-) = 0.5256,$
$S(M^3, M^-) = 0.5531,$   $S(M^4, M^-) = 0.5339$

Using the formula (7) find the values of $f(t_i)$, we get

$f(t_1) = 0.7034,$   $f(t_2) = 0.6541,$
$f(t_3) = 0.6040$   $f(t_4) = 0.6166$

Now, $f(t_1) > f(t_2) > f(t_4) > f(t_3)$ such that

$$t_1 > t_2 > t_4 > t_3.$$

Hence $t_1$ place the highest position, while $t_3$ place the last position, finally $t_2$ and $t_4$ places the centre position orders.

In the same way, to calculate the position outcomes of the values $\gamma = 0, 0.3, 0.7\ and\ 1.0$ by using the above working procedure in the following tables 3 and 4.

Table 1.The ranking order of the replacements for distinct values of $\gamma$ using $Xu$'s technique and working procedure

| $\gamma$ | $C$ | $S(M^i, M^+)$ | $S(M^i, M^-)$ |
|---|---|---|---|
| 0 | $C_1 = (0.3616, 0.3821, 0.2564)$<br>$C_2 = (0.3041, 0.4375, 0.2584)$<br>$C_3 = (0.3062, 0.3626, 0.3734)$ | $(0.3701, 0.3558, 0.3463, 0.3400)$ | $(0.4955, 0.5125, 0.5416, 0.5242)$ |
| 0.3 | $C_1 = (0.3255, 0.3764, 0.2667)$<br>$C_2 = (0.2971, 0.4670, 0.2359)$<br>$C_3 = (0.3090, 0.3595, 0.3905)$ | $(0.3612, 0.3464, 0.3384, 0.3310)$ | $(0.5053, 0.5228, 0.5508, 0.5341)$ |
| 0.5 | $C_1 = (0.3616, 0.3821, 0.2564)$<br>$C_2 = (0.2925, 0.4866, 0.2692)$<br>$C_3 = (0.3062, 0.3626, 0.3734)$ | $(0.3567, 0.3438, 0.3341, 0.3292)$ | $(0.5071, 0.5256, 0.5531, 0.5339)$ |
| 0.7 | $C_1 = (0.3527, 0.3878, 0.2461)$<br>$C_2 = (0.2878, 0.4662, 0.2060)$<br>$C_3 = (0.3034, 0.3657, 0.3563)$ | $(0.3577, 0.3430, 0.3344, 0.3282)$ | $(0.5114, 0.5285, 0.5587, 0.5363)$ |
| 1.0 | $C_1 = (0.3730, 0.3963, 0.2307)$<br>$C_2 = (0.2808, 0.5357, 0.1835)$<br>$C_3 = (0.2991, 0.3703, 0.3306)$ | $(0.3418, 0.3261, 0.3204, 0.3120)$ | $(0.5263, 0.5425, 0.5731, 0.5498)$ |

Table 2.The ranking order of the replacements by using $Xu$'s technique and working procedure.

| $\gamma$ | $f(t_1)$ | $f(t_2)$ | $f(t_3)$ | $f(t_4)$ | Ranking |
|---|---|---|---|---|---|
| 0 | 0.4276 | 0.4098 | 0.3900 | 0.3934 | $t_1 > t_2 > t_4 > t_3$ |
| 0.3 | 0.4169 | 0.3985 | 0.3805 | 0.3826 | $t_1 > t_2 > t_4 > t_3$ |
| 0.5 | 0.7034 | 0.6541 | 0.6040 | 0.6166 | $t_1 > t_2 > t_4 > t_3$ |
| 0.7 | 0.4115 | 0.3936 | 0.3745 | 0.3796 | $t_1 > t_2 > t_4 > t_3$ |
| 1.0 | 0.3937 | 0.3754 | 0.3586 | 0.3620 | $t_1 > t_2 > t_4 > t_3$ |

According to the working procedure and XU's technique, by substituting the values of $\gamma = 0, 0.3, 0.5, 0.7\ and\ 1.0$, we get the same results for all the values.

Therefore, $t_1 > t_2 > t_4 > t_3$

Hence $t_1$ place the highest position, while $t_3$ place the last position, finally $t_2$ and $t_4$ places the centre position orders and which is mentioned in the above tables

*(ii) The Laplacian energy of HFG*

By the definition of LE of HFG is

$$LE = \left| \kappa_i - \frac{2 \sum_{1 \leq i \leq j \leq n} \mu(t_i, t_j)}{n} \right|$$

Where the Laplacian matrix of HFG is $L = D - A$, here D is the degree matrix and A is the adjacency matrix.

The Laplacian energy of the Laplacian matrices are defined as below

$LE(M^{(1)}) = (2.1000, 2.1639, 1.3000)$

$LE(M^{(2)}) = (1.8000, 2.7000, 0.9290)$

$LE(M^{(3)}) = (1.6000, 2.0000, 2.4000)$.

The scores of each expert can calculate as:

$$C_b^1 = \left((C_\mu)_i, (C_\gamma)_i, (C_\beta)_i\right) = \left[\frac{LE((D_\mu)_i)}{\sum_{r=1}^{l} LE((D_\mu)_r)}, \frac{LE((D_\gamma)_i)}{\sum_{r=1}^{l} LE((D_\gamma)_r)}, \frac{LE((D_\beta)_i)}{\sum_{l=1}^{l} LE((D_\beta)_r)}\right], i = ,2,3,\ldots,n$$

The calculation of every expert's scores is as follows:

$$C_1^1 = [0.3818, 0.3153, 0.2808],$$
$$C_2^1 = [0.3273, 0.3934, 0.2007]$$
$$C_3^1 = [0.2909, 0.2914, 0.5185]$$

By evaluating the SMs $S(M^{(b)}, M^{(d)})$ between $M^{(b)}$ and $M^{(d)}$ for every $b \neq d$:

$$S(M^{(1)}, M^{(2)}) = 1.9856,$$
$$S(M^{(2)}, M^{(3)}) = 2.0570,$$
$$S(M^{(1)}, M^{(3)}) = 1.9155$$

The average SD $S(M^{(k)})$ of $M^{(k)}$ is obtained as below,

$$S(M^{(k)}) = 1.9506,$$
$$S(M^{(k)}) = 2.0213,$$
$$S(M^{(k)}) = 1.9863.$$

To find the scores $C_b^a$ using $S(M^{(1)})$, $S(M^{(2)})$ and $S(M^{(3)})$, we get

$$C^b = (0.3274, 0.3392, 0.3334)$$

By evaluating the values of the objective scoresing vector using the value $\eta = 0.5$ we get

$$C_1^2 = (0.3546, 0.3273, 0.3071),$$
$$C_2^2 = (0.3274, 0.3663, 0.2671)$$
$$C_3^2 = (0.3092, 0.3153, 0.4260)$$

Now determine the values of the objective and subjective scores $C_b$ using the formula (9) and $\gamma = 0.5$, , we have

$$C_1 = (0.3676, 0.3213, 0.2940),$$
$$C_2 = (0.3274, 0.3799, 0.2339)$$
$$C_3 = (0.3001, 0.3034, 0.4723)$$

To determine the hesitancy preference relation $M = (r_{ij})_{n \times n}$, we get

$$M = \begin{bmatrix} (0,0,0) & (0.3053, 0.3542, 0.3005) & (0.3680, 0.3317, 0.1528) & (0.2331, 0.4095, 0.2945) \\ (0.3053, 0.3542, 0.3005) & (0,0,0) & (0.3053, 0.4077, 0.1294) & (0.2030, 0.4113, 0.3183) \\ (0.3680, 0.3317, 0.1528) & (0.3053, 0.4077, 0.1294) & (0,0,0) & (0.2631, 0.4340, 0.1528) \\ (0.2331, 0.4095, 0.2945) & (0.2030, 0.4113, 0.3183) & (0.2631, 0.4340, 0.1528) & (0,0,0) \end{bmatrix}$$

By evaluating the 'SMs' $S(M^i, M^+)$ between $M^i$ and $M^+$, we have
$$S(M^1, M^+) = 0.3953, \quad S(M^2, M^+) = 0.3824,$$
$$S(M^3, M^+) = 0.3712, \quad S(M^4, M^+) = 0.3671$$

By evaluating the 'SMs' $S(M^i, M^-)$ between $M^i$ and $M^+$, we have
$$S(M^1, M^-) = 0.4782, \quad S(M^2, M^-) = 0.4928,$$
$$S(M^3, M^-) = 0.5239, \quad S(M^4, M^-) = 0.5062$$

Using the formula (7) find the values of $f(x_i)$, we get
$$f(t_1) = 0.4526, \quad f(t_2) = 0.4369,$$
$$f(t_3) = 0.4147 \quad f(t_4) = 0.4203$$

Then, $f(t_1) > f(t_2) > f(t_4) > f(t_3)$.

Hence $t_1 > t_2 > t_4 > t_3$.

According to the above working procedure, we have

Hence $x_1$ place the highest position, while $x_3$ place the last position, finally $x_2$ and $x_4$ places the centre position order.

In the same way, to calculate the position outcomes of the values $\gamma = 0, 0.3, 0.7\ and\ 1.0$ by using the above working procedure in the following tables 3 and 4.

Table 3. The ranking order of the replacements for distinct values of $\gamma$ using $Xu's$ technique and working procedure

| $\gamma$ | $C$ | $S(M^i, M^+)$ | $S(M^i, M^-)$ |
|---|---|---|---|
| 0 | $C_1 = (0.3546, 0.3273, 0.3071)$ $C_2 = (0.3274, 0.3663, 0.2671)$ $C_3 = (0.3092, 0.3153, 0.4260)$ | $(0.3949, 0.3822, 0.3711, 0.3671)$ | $(0.4794, 0.4944, 0.5136, 0.5060)$ |
| 0.3 | $C_1 = (0.3628, 0.3237, 0.2992)$ $C_2 = (0.3274, 0.3744, 0.2472)$ $C_3 = (0.3037, 0.3081, 0.4538)$ | $(0.3938, 0.3823, 0.3712, 0.3657)$ | $(0.4834, 0.4934, 0.5234, 0.5109)$ |
| 0.5 | $C_1 = (0.3676, 0.3213, 0.2940)$ $C_2 = (0.3274, 0.3799, 0.2339)$ $C_3 = (0.3001, 0.3034, 0.4723)$ | $(0.3953, 0.3824, 0.3712, 0.3671)$ | $(0.4782, 0.4928, 0.5239, 0.5062)$ |
| 0.7 | $C_1 = (0.3736, 0.3189, 0.2887)$ $C_2 = (0.3273, 0.3853, 0.2206)$ $C_3 = (0.2964, 0.2986, 0.4903)$ | $(0.3955, 0.3826, 0.3713, 0.3672)$ | $(0.4777, 0.4920, 0.5244, 0.5060)$ |
| 1.0 | $C_1 = (0.3818, 0.3153, 0.2808)$ $C_2 = (0.3273, 0.3934, 0.2007)$ $C_3 = (0.2909, 0.2914, 0.5185)$ | $(0.3956, 0.3824, 0.3714, 0.3671)$ | $(0.4773, 0.4915, 0.5251, 0.5063)$ |

Table 4. The ranking order of the replacements by using $Xu's$ technique and working procedure.

| $\gamma$ | $f(t_1)$ | $f(t_2)$ | $f(t_3)$ | $f(t_4)$ | Ranking |
|---|---|---|---|---|---|
| 0 | 0.4520 | 0.4360 | 0.4194 | 0.4204 | $t_1 > t_2 > t_4 > t_3$ |
| 0.3 | 0.4489 | 0.4366 | 0.4149 | 0.4172 | $t_1 > t_2 > t_4 > t_3$ |
| 0.5 | 0.4526 | 0.4369 | 0.4147 | 0.4203 | $t_1 > t_2 > t_4 > t_3$ |
| 0.7 | 0.4530 | 0.4375 | 0.4146 | 0.4205 | $t_1 > t_2 > t_4 > t_3$ |
| 1.0 | 0.4532 | 0.4376 | 0.4142 | 0.4203 | $t_1 > t_2 > t_4 > t_3$ |

According to the working procedure and XU's technique, by substituting the values of $\gamma = 0, 0.3, 0.5, 0.7 \text{ and } 1.0$, we get the same results for all the values.

Therefore, $t_1 > t_2 > t_4 > t_3$.

Hence $t_1$ place the highest position, while $x_3$ place the last position, finally $t_2$ and $t_4$ places the centre position orders and which is mentioned in the above tables.

## 3. Conclusion

In this article, we extract evidence from the experts' implemented judgments, namely different HFPRs to the replacements, and change it into the experts' actual scores using working Procedure.We introduced a better technique for evaluating the similar reputational scores of experts by estimating the ambiguous information of HFPRs and the mean similarity grades between one HFPR and another.Recently, many variance and similitude estimation strategies have been applied to GDM problems based on HFPR and also we extract evidence from the experts' implemented judgments, namely different HFPRs to the replacements, and change it into the experts' actual scores using working Procedure.additionally, we merged the different HFPrss into a merged HFPR using an hesitancy fuzzy scoring average operator and defined a relevant similarity approach to derive the emergencies of replacements from the merged HFPR. Furthermore, examples are provided that enhances the accuracy of the similarity measurements. The Ranking order of the energy and Laplacian energy outcomes are same.

In the future, the application of the proposed "cosine similarity measures" of PFSs needs to be explored in complex decision making, [21–28] risk analysis, and many other fields under uncertain environments [12,29–32] and [34, 35].